\documentclass{article}

\usepackage{xcolor}
\usepackage[final]{corl_2026} 
\usepackage{graphicx}
\usepackage{wrapfig}
\usepackage{amsmath}
\usepackage{amssymb}
\usepackage{booktabs}
\usepackage{nicematrix}
\usepackage{booktabs,makecell,multirow}
\usepackage{adjustbox}
\usepackage{xspace}
\usepackage{enumitem}
\usepackage[ruled,vlined]{algorithm2e}

\newcommand\blfootnote[1]{%
  \begingroup
  \renewcommand\thefootnote{}\footnote{#1}%
  \addtocounter{footnote}{-1}%
  \endgroup
}

\newcommand{\coolname}{\texttt{LateMatch}\xspace}
\newcommand{\estore}{\texttt{E-SToRE}\xspace}
\newcommand{\estoren}{E-SToRE\xspace} 

\newcommand{\lightglue}{LightGlue\xspace}
\newcommand{\mambaglue}{MambaGlue\xspace}

\newcommand{\lightgluelate}[1][]{\texttt{LightGlue-late#1}\xspace}
\newcommand{\mambagluelate}[1][]{\texttt{MambaGlue-late#1}\xspace}
\newcommand{\lightgluelite}[1][]{\texttt{LightGlue-lite#1}\xspace}
\newcommand{\mambagluelite}[1][]{\texttt{MambaGlue-lite#1}\xspace}

\newcommand{\lglate}[1][]{LG-late#1\xspace}
\newcommand{\lglite}[1][]{LG-lite#1\xspace}

\title{\coolname: Breaking the Interaction Bottleneck via Cacheable  Matching for Efficient Visual Localization}

\author{
  Mohammad Omama\\
  The University of Texas \\ at Austin \\ 
  \texttt{mohd.omama@utexas.edu} \\
  \And
  Aeyan Ashraf \\
  University of Massachusetts \\ Amherst  \\
  \texttt{aeyanashraf@gmail.com} \\
  \And
  Sandeep Chinchali \\
  The University of Texas \\ at Austin \\
  \texttt{sandeepc@utexas.edu} \\
}

\begin{document}

\maketitle
\blfootnote{Code: \url{https://github.com/UTAustin-SwarmLab/LateMatch}}


\begin{abstract}

Visual localization is critical for embodied agents, but deploying state-of-the-art deep feature matchers on edge devices is fundamentally limited by their massive computational cost. Standard deep matchers tightly interleave intra-image context building and inter-image matching at every layer, creating a severe ``interaction bottleneck.'' Localizing a single query against $K$ database images forces the system to execute the full network $K$ independent times with no capacity for caching. To break this bottleneck, we advocate for a paradigm shift toward \textit{cacheable image matching}. We realize this paradigm with \textbf{\coolname}, a versatile approach that restructures existing matchers by deferring inter-image interaction to their final layers. By isolating intra-image processing into offline-cacheable Context Blocks, only a small set of Interaction Blocks must be executed online. To further optimize \coolname, we introduce \textbf{\estore} (Entropy-guided Spatial Token Reduction and Expansion). \estore leverages the decreasing Shannon entropy of the late Interaction Blocks to dynamically reduce token counts just prior to cross-image interaction. Evaluated across HPatches, ScanNet, MegaDepth, ETH-3D, and Aachen, \coolname reduces effective computational cost by over 2$\times$ compared to standard deep matchers, while \estore yields an additional 10\% to 14\% savings with negligible degradation in accuracy.
\end{abstract}


\section{Introduction}\label{sec:intro}

Precise and real-time visual localization is a fundamental capability for embodied agents, spanning from autonomous robots to augmented reality (AR) systems. Today, these systems capture rich sensory data that must be aligned against pre-mapped environments to estimate absolute camera pose. Deploying these localization pipelines on resource-constrained edge devices, such as smart glasses or drones, requires algorithms that operate under strict compute, latency, and battery constraints. 

\begin{figure}[t]
    \centering
    \includegraphics[width=\linewidth]{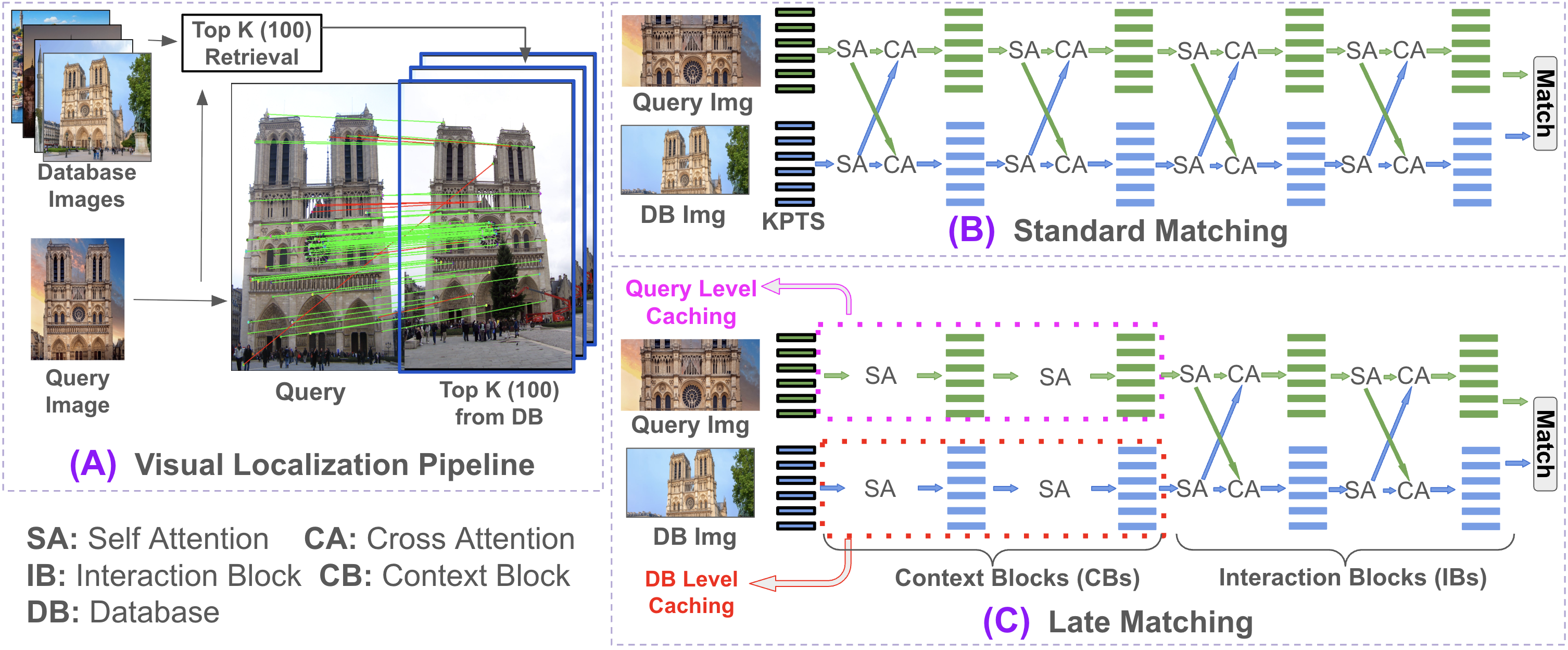}
    \caption{
    \small
    \textbf{Rethinking visual localization with cacheable image matching via \coolname.} \textbf{(A)} Standard visual localization pipelines require matching a single query image against Top-$K$ (e.g., $K=100$) retrieved database images. \textbf{(B)} Traditional matching architectures interleave intra-image context building and inter-image cross-attention across all layers, preventing any caching and causing a severe computational bottleneck. \textbf{(C)} Our proposed \coolname architecture breaks this bottleneck by deferring cross-image interaction to the final layers, decoupling the network into offline-cacheable Context Blocks (CBs) and online Interaction Blocks (IBs), reducing the effective compute cost by over 2$\times$ without sacrificing much on performance.}
    \label{fig:teaser}
\end{figure}

Despite the critical need for edge-deployable localization, a key open problem is the massive computational cost of state-of-the-art deep feature matchers. Visual localization inherently requires retrieving $K$ candidate database images and matching a single query image against each of them as shown in ~\autoref{fig:teaser}\textbf{(A)}. Standard deep matchers treat this $1$-to-$K$ matching task as $K$ independent pairwise operations. Because these models interleave self-attention (or its variants) and cross-attention at every layer, they must feed each query-database pair through the entire network to compute a match. This creates a severe ``interaction bottleneck'' that renders full-capacity models unusable for real-time edge deployment. While existing efficiency strategies rely on reduced-capacity architectures, these methods severely compromise on performance to achieve latency reductions.

Our key observation is that state-of-the-art deep matchers fail to scale efficiently because they tightly interleave intra-image context building (e.g., self-attention or variants) and inter-image matching (e.g., cross-attention or variants) at every layer. This forces the system to treat the matching of $K$ images in the visual localization pipeline as $K$ independent pairwise operations with no capacity for caching (pre-computation). To address this, our key technical insight is to internally restructure the deep matcher by deferring inter-image interaction to the final layers. This fundamentally divides the matcher into two parts: a set of cacheable Context Blocks (CBs) that build intra-image context, and a set of Interaction Blocks (IBs) at the very end that extract information across images while further refining internal representations, as shown in~\autoref{fig:teaser}\textbf{(C)}. This approach is fundamentally different from simply using a lightweight matcher with fewer layers. \textit{By delaying but retaining} the interaction, training signals pass fully through the Context Blocks, enabling them to build highly discriminative representations that can be matched accurately with very few interaction steps. 

This \coolname structure enables a new paradigm: \textit{cacheable image matching}. First, it allows us to pre-compute and cache the output of the Context Blocks for all database images offline. Second, during online inference, the query image passes through the Context Blocks exactly once. Only the final Interaction Blocks must be executed $K$ times against the retrieved database images. To quantify this, we introduce the concept of Effective-GigaFLOPs (E-GFLOPs or E-GF). Because $K$ is large in standard retrieval pipelines, the amortized cost of the query's Context Blocks becomes negligible, meaning the E-GFLOPs per image pair approaches only the cost of the Interaction Blocks (detailed in Sec.~\ref{sec: e-gflops}). For instance, a 9-layer late matcher  with 5 Context Blocks and 4 Interaction Blocks reduces the E-GFLOPs by up to 2.25$\times$ compared to a standard 9-layer deep matcher.

\begin{figure}[t]
    \centering
    \includegraphics[width=\linewidth]{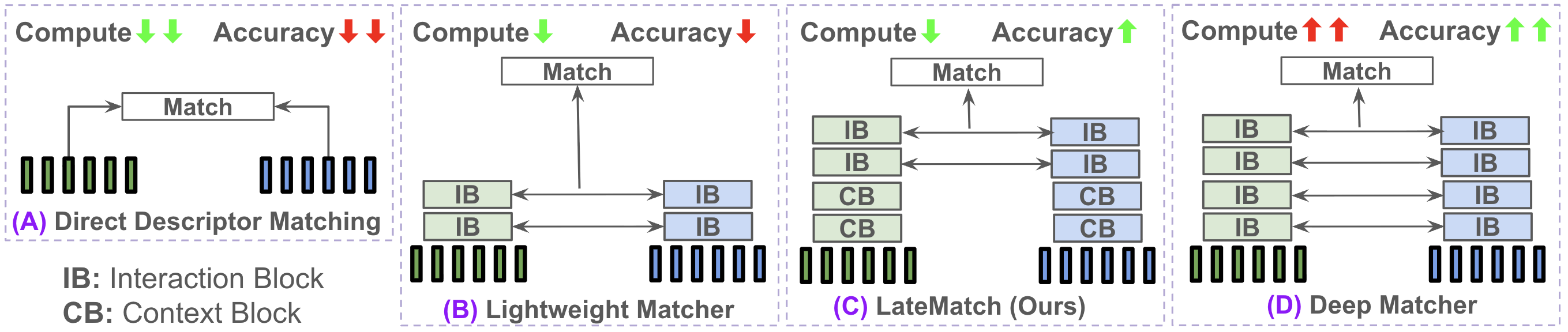}
    \caption{
        \small
        \textbf{Strategic positioning of \coolname in the feature matching landscape.} While direct nearest-neighbor matching (\textbf{A}) is extremely fast but highly inaccurate, and lightweight matchers (\textbf{B}) are fast but nowhere close to the accuracy of heavy, computationally prohibitive deep matchers (\textbf{D}), our proposed \coolname paradigm (\textbf{C}) achieves the high performance of full-capacity models with the online computational footprint of lightweight matchers.
    }
    \label{fig:match_comp}
\end{figure}

The \coolname paradigm has a unique and strategic positioning within the feature matching landscape, as shown in ~\autoref{fig:match_comp}. Direct nearest-neighbor matching is fast but yields poor accuracy. Lightweight matchers reduce compute but suffer significant performance drops. Standard deep matchers provide high accuracy but incur prohibitive computational costs. Our approach bridges this gap, delivering the high performance of deep matchers with the computational efficiency of lightweight alternatives.

We further analyze \coolname from an information-theoretic perspective. We observe that the Shannon entropy of the interaction layers starts high and decreases with each successive Interaction Block. This indicates that early interactions attend broadly to many keypoints across the images, while later interactions focus precisely on a few. To exploit this behavior, we introduce Entropy-guided Spatial Token Reduction and Expansion (\estore). \estore merges the $r$ spatially closest token pairs before cross-attention, shortening the sequence by $r$ tokens. It then computes the interaction and unmerges the tokens for the subsequent layer. The reduction budget $r$ is initialized with a high value in the early Interaction Blocks and progressively decreases in deeper blocks. \estore provides an additional $10\%$ to $14\%$ of computational reduction on top of \coolname.

Our contributions are four-fold:
\begin{enumerate}
     \item We introduce \coolname, a novel paradigm that breaks the interaction bottleneck in $1$-to-$K$ visual localization by deferring inter-image interaction to the final layers, thereby enabling \textit{cacheable image matching}.
    \item We introduce Effective-GigaFLOPs (E-GFLOPs) as a formal metric to quantify the computational footprint of $1$-to-$K$ image matching.
    \item We demonstrate the generality of \coolname by applying it to both attention-based (\lightglue~\citep{lightglue2023}) and state-space-based (\mambaglue~\citep{ryoo2025mambaglue}) matchers. This results in the \lightgluelate and \mambagluelate model families, which reduce effective compute by over 2$\times$ while remaining comparable in performance to standard full-capacity models.
    \item We propose \estore (Entropy-guided Spatial Token Reduction and Expansion), a dynamic mechanism that leverages the decreasing Shannon entropy of cross-attention in the  Interaction Blocks of \coolname to reduce and restore token counts. \estore provides an additional $10\%$ to $14\%$ of computational reduction on top of \coolname.
\end{enumerate}


\section{Related Work}\label{sec:related}

\textbf{Visual Localization.} Visual localization is generally classified into structure-based and image-based paradigms. \emph{Structure-based methodologies} localize the camera by directly matching query image features to a 3D point cloud reconstructed via Structure-from-Motion (SfM) \citep{svarm2016city, toft2018semantic, sattler2016efficient, liu2017efficient, taira2018inloc}. In contrast, \emph{image-based methodologies} provide a highly robust and popular alternative through a coarse-to-fine hierarchical approach. Standard pipelines within this category \citep{Sattler_2017_large_scale_3d, sattler2018aachen, hloc} first employ visual place recognition models \citep{NetVlad, CosPlace, Izquierdo_CVPR_2024_SALAD, MegaLoc, Omama2024exploiting, omama2026ssr, li2024any2any} to retrieve a subset of visually similar database images, which are subsequently subjected to image matching to estimate the final pose. 

\textbf{Learned Detectors and Descriptors.}
Learned local features \citep{superpoint, d2net, luo2019contextdesc, tian2019sosnet, r2d2, tyszkiewicz2020disk, gleize2023silk, Zhao2022ALIKE, potje2024xfeat, omama2026asymloc} have driven substantial progress in image matching in recent years. These methods extract keypoints and descriptors from an image, which are matched using basic heuristics like nearest neighbors. Prominent examples include SuperPoint \citep{superpoint}, which utilizes self-supervised training on synthetic shapes and homographies. DISK \citep{tyszkiewicz2020disk} optimizes directly for downstream matching performance via reinforcement learning, while architectures like ALIKE \citep{Zhao2022ALIKE} and XFEAT \citep{potje2024xfeat} prioritize highly efficient, lightweight feature extraction. Although highly compute-efficient, these descriptors often fall significantly behind deep matchers in terms of matching accuracy.

\textbf{Deep Sparse Matchers.}  
Learned deep sparse matchers \citep{superglue, seededgraphmatching, lightglue2023, omniglue, ryoo2025mambaglue} have largely superseded nearest-neighbor heuristics by leveraging neural networks to model keypoint matching on top of existing detectors and descriptors.
While early state-of-the-art models like SuperGlue~\citep{superglue} relied on heavy graph neural networks, recent architectures prioritize efficiency. LightGlue~\citep{lightglue2023}, for instance, achieves high accuracy efficiently using interleaved self- and cross-attention transformer blocks, while MambaGlue~\citep{ryoo2025mambaglue} further improves performance by integrating state-space Mamba models~\citep{mamba, mamba2, mamba3}. However, despite these architectural optimizations, these efficient matchers still execute $1$-to-$K$ visual localization as $K$ completely independent pairwise operations with no possibility of pre-computation (or caching), creating a severe computational bottleneck. 

\textbf{End-to-end Dense Methods.} End-to-end dense methods, like RoMa \citep{densse_roma}, LoFTR \citep{loftr}, and others \citep{dense_minima_roma, dense_se2_loftr, efloftr, dense_aspanformer, dense_matchformer, dense_jung2025edm, Chen_2025_RDD}, achieve remarkable robustness by jointly processing both keypoint extraction and matching through monolithic network architectures. However, this joint processing incurs a massive baseline computational footprint. While the \coolname paradigm of deferring interaction could theoretically be extended to these methods, their overall parameter counts remain fundamentally prohibitive for the resource-constrained edge devices. Consequently, extending cacheable matching to dense frameworks is left as an avenue for future work.

\textbf{Late Interaction.}
Late interaction has been studied mainly in retrieval. ColBERT~\citep{colbert} independently encodes queries and documents before applying token-level interaction for passage scoring; ColBERTv2~\citep{colbertv2} improves this design with residual compression and denoised supervision; and PLAID~\citep{plaid} accelerates late-interaction retrieval using centroid interaction and pruning. Similar ideas have also appeared in multimodal retrieval, such as FILIP~\citep{filip} for fine-grained image-text alignment and Video-ColBERT~\citep{videocolbert} for text-to-video retrieval. These works show that late interaction can preserve fine-grained similarity while enabling pre-computation (caching). However, feature matching is different from retrieval because it must output geometrically consistent keypoint correspondences, not only a relevance score. Our work is the first to adapt late interaction to deep keypoint matching and study it from an information-theoretic lens to further optimize it.


\section{Problem Statement}
\label{sec:problem_statement}

We will now formalize the standard visual localization pipeline and mathematically define the computational bottleneck that prevents state-of-the-art deep matchers from scaling efficiently.

\subsection{Visual Localization as $1$-to-$K$ Matching}\label{sec:one_to_k}
The core task in visual localization is to estimate the camera pose of a query image $I_q$ within a pre-mapped environment. While the global map contains a massive collection of reference images, a retrieval step first extracts a localized subset of $K$ candidates to serve as the active database for matching (see ~\autoref{fig:teaser}\textbf{A}). We denote this retrieved database as $\mathcal{I}_{db} = \{{}^1I_{db}, {}^2I_{db}, \dots, {}^KI_{db}\}$.\\
The standard matching pipeline first utilizes a local feature extractor $\phi$ (e.g., SuperPoint~\citep{superpoint}) to process each image independently. This yields a set of $N$ spatial keypoints $P \in \mathbb{R}^{N \times 2}$ and their corresponding $d$-dimensional descriptors $D \in \mathbb{R}^{N \times d}$, where $N$ depends on the image. For any image $I$, the extraction is defined as $(P, D) = \phi(I)$. 
Once features are extracted, a deep matcher $f_\theta$ is deployed to find accurate correspondences between the query features $(P_q, D_q)$ and each of the $k \in \{1,\dots,K\}$ database features $({}^kP_{db}, {}^kD_{{db}})$, which contain $N_q$ and ${}^kN_{db}$ keypoints, respectively.

\subsection{The Interaction Bottleneck}
Let $f_\theta$ be a standard deep matcher, such as LightGlue or MambaGlue, consisting of $L$ interleaved attention layers (or variants). At any given layer $l$, the matcher updates the feature representations $F_q^l$ and $F_{db}^l$ by computing both intra-image self-attention (or variant as in MambaGlue) and inter-image cross-attention. Mathematically, the forward pass for a single layer is:
\begin{equation}
F_q^{l}, F_{db}^{l} = \mathrm{Layer}^l(F_q^{l-1}, F_{db}^{l-1}).
\end{equation}

Here and throughout, we suppress the database index $k$ when describing a single query-database pair, writing $F_{db}^{l}$ for ${}^kF_{db}^{l}$. Each layer additionally receives the keypoint positions $P_q$ and $P_{db}$, which we omit from the notation for brevity. Note that $F_{q}^{0}  = D_{q}$, $F_{db}^{0}  = D_{db}$. Crucially, because the database features $F_{db}^{l}$ are generated using cross-attention from the query features $F_q^{l-1}$ at every layer, the intermediate database representations become strictly conditioned on that specific query image and vice versa. This structural entanglement makes it mathematically impossible to pre-compute or cache the database features offline. 
Consequently, to localize a single query against $K$ database images, the entire $L$-layer network must be executed online $K$ separate times. If $\mathcal{C}(\cdot)$ is the per-pair forward-pass cost, the total  cost of the standard matcher ($\mathcal{C}_{\mathrm{total}}^{\mathrm{std}}$) becomes:
{
\small
\begin{equation}
\mathcal{C}_{\mathrm{total}}^{\mathrm{std}} = K \cdot \mathcal{C}(f_\theta).
\end{equation}
}
\begin{wraptable}{r}{0.56\textwidth}
\vspace{-1em}
\centering
\footnotesize
\begin{tabular}{l r r r}
\toprule
\textbf{Stage (online)} & \textbf{GF} & \makecell{\textbf{Lat.} \\ (ms)} & \makecell{\textbf{Lat.} \\ (\%)} \\
\midrule
SuperPoint (query, $\times 1$) & 0.3k & 7.2 & 0.4 \\
Matching ($\times K{=}50$) & \textbf{27.9k} & \textbf{1558.3} & \textbf{96.2} \\
PnP + RANSAC ($\times 1$) & $\sim$0.025 & 54.3 & 3.4 \\
\midrule
Total & 28.3k & 1619.8 & 100.0 \\
\bottomrule
\end{tabular}
\caption{
    \footnotesize
    \textbf{Matching dominates $1$-to-$K$ localization cost.} Stage-wise cost of a standard \lightglue pipeline on Aachen ($K{=}50$, desktop GPU). GigaFLOPs (GF) is the total online cost, not the per-pair cost of Sec.~\ref{sec: e-gflops}.
}
\label{tab:lat_breakdown}
\end{wraptable}

\noindent For real-time applications, executing a full $L$-layer matcher $K$ separate times becomes computationally prohibitive. \autoref{tab:lat_breakdown} quantifies this on the Aachen  dataset~\citep{sattler2018aachen} with $K{=}50$ (see Sec.~\ref{sec:exp_setup}). Feature extraction and pose solving each run once per query, but matching runs $K$ times and consumes $96.2\%$ of the end-to-end latency. Matching is therefore the dominant online cost in $1$-to-$K$ localization, and it is exactly the stage that standard matchers cannot cache.
\vspace{-1em}
\section{Methodology}
\label{sec:method}

To break the interaction bottleneck, we propose \coolname. We internally restructure the deep matcher by decoupling intra-image context building from inter-image matching.

\subsection{Decoupled Architecture}
We divide the $L$-layer architecture into two distinct functional stages: $L_{CB}$ Context Blocks (CBs) and $L_{IB}$ Interaction Blocks (IBs), such that $L = L_{CB} + L_{IB}$. \\
\textbf{Context Blocks (CBs):} The first $L_{CB}$ layers are dedicated strictly to intra-image processing (e.g., self-attention, Mamba, or variants). Devoid of inter-image cross-attention, these blocks produce representations that are completely independent of any paired image, making them fully cacheable. \\
\textbf{Interaction Blocks (IBs):} The final $L_{IB}$ layers retain the full architectural structure, interleaving both intra-image processing and inter-image cross-attention to perform the actual matching. During online inference, the query image passes through the Context Blocks exactly once to yield $F_q^{L_{CB}}$. Subsequently, only the Interaction Blocks are executed $K$ times to match $F_q^{L_{CB}}$ against the $K$ pre-cached database representations ${}^kF_{db}^{L_{CB}}$.

\subsection{Model Instantiations}\label{sec:model_ins}
To demonstrate generalizability, we instantiate \coolname into two different model families: 

\lightgluelate: This family is built upon the attention-based \lightglue~\citep{lightglue2023} matcher. Standard \lightglue consists of 9 layers, each tightly interleaving self-attention for intra-image processing and cross-attention for inter-image matching. We restructure this by stripping the cross-attention from the early layers, transforming them into pure Context Blocks (CBs) that solely execute intra-image self-attention. The final layers, however, retain the full original structure, interleaving both intra-image self-attention and inter-image cross-attention, and serve as the Interaction Blocks (IBs). We also introduce the notation \lightgluelate{$x$}, where $x$ denotes the exact number of late IBs. Consequently, a \lightgluelate{$x$} model consists of $(9-x)$ cacheable CBs followed by $x$ IBs.

\mambagluelate: This family is built upon the \mambaglue~\citep{ryoo2025mambaglue} matcher. Standard \mambaglue also spans 9 layers, utilizing a mixture of Mamba and self-attention for intra-image processing, followed by cross-attention for inter-image matching at every layer. Applying our paradigm, we remove the cross-attention from the early layers to form Context Blocks (CBs) that rely solely on Mamba and self-attention. The final Interaction Blocks (IBs) maintain the complete interleaved design, executing both the intra-image and the inter-image processing. Following the same notation, a \mambagluelate{$x$} model comprises $(9-x)$ cacheable CBs and $x$ IBs.

\subsection{Effective-GigaFLOPs (E-GFLOPs)}\label{sec: e-gflops}
To formally quantify the computational savings of this paradigm, we introduce the concept of Effective-GigaFLOPs (E-GFLOPs or E-GF). The total online cost to localize a query against $K$ cached database images is the cost of processing the query through the Context Blocks plus the cost of running the Interaction Blocks $K$ times:
{
\small
\begin{equation}
    \mathcal{C}_{\mathrm{total}}^{\mathrm{late}} = \mathcal{C}_{CB}(I_q) + K \cdot \mathcal{C}_{IB}(I_q, I_{db}).
\end{equation}
}
E-GFLOPs defines the amortized computational footprint per image pair. By dividing the total cost by $K$, we get:
{
\small
\begin{equation}
\mathrm{E\text{-}GFLOPs} = \frac{\mathcal{C}_{CB}(I_q)}{K} + \mathcal{C}_{IB}(I_q, I_{db}).
\end{equation}
}
\begin{wrapfigure}[14]{r}{0.38\textwidth}
    \vspace{-1.4em}
    \centering
    \includegraphics[width=\linewidth]{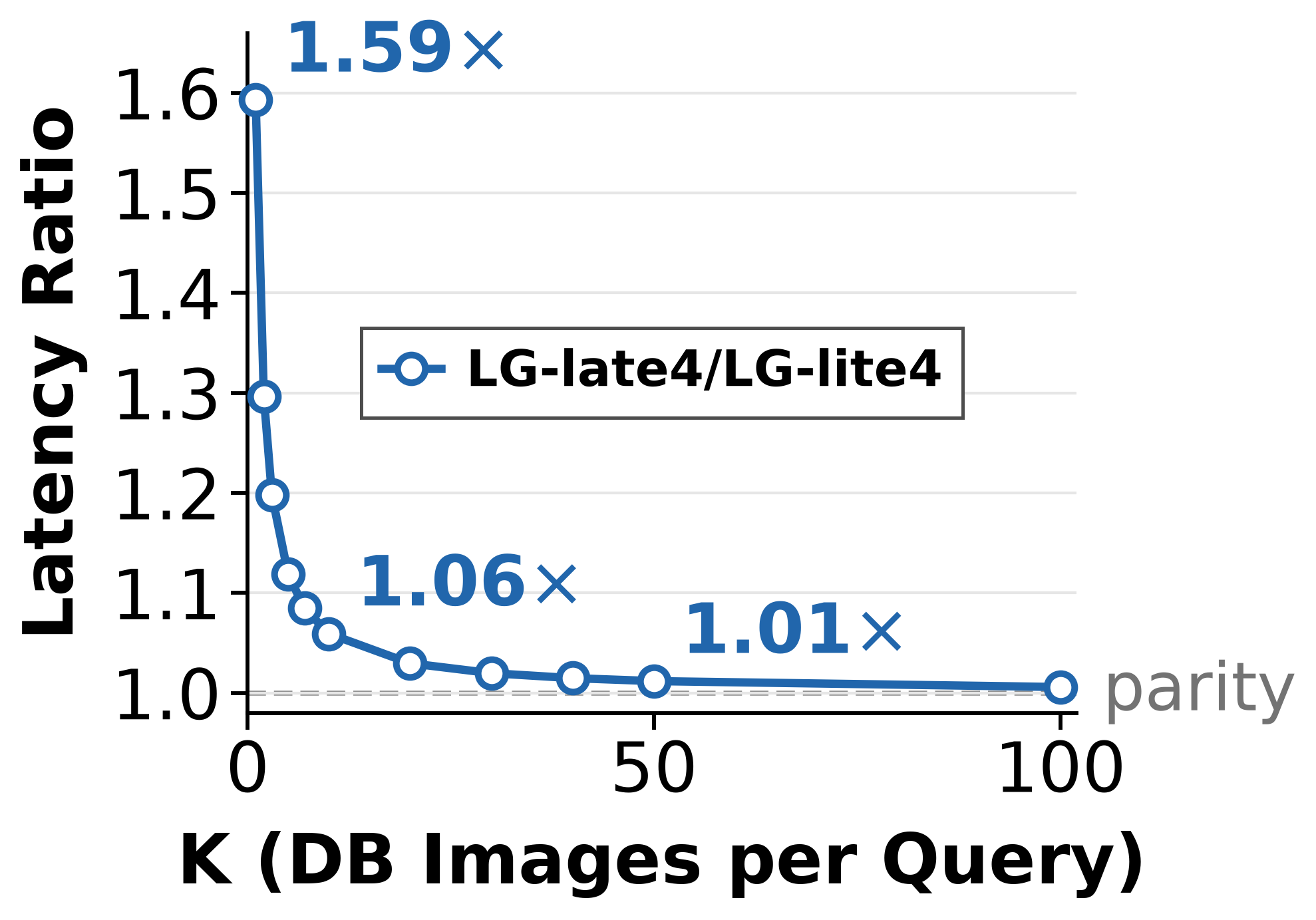}
    \caption{
    \footnotesize
    \textbf{Context Block cost amortizes by $K{\approx}10$.} \lglate{4} latency relative to \lglite{4} on Aachen.}
    \label{fig:break_even_k}
    \vspace{-1em}
\end{wrapfigure}

In standard retrieval pipelines where $K$ is large (typically $K \ge 50$), the amortized cost of the query's Context Blocks is marginalized by the interaction cost, such that $\frac{\mathcal{C}_{CB}(I_q)}{K} \ll \mathcal{C}_{IB}(I_q, I_{db})$. Hence, the effective compute per image pair reduces to the cost of the Interaction Blocks:
{
\small
\begin{equation}
\mathrm{E\text{-}GFLOPs} \approx \mathcal{C}_{IB}(I_q, I_{db}).
\end{equation}
}
\autoref{fig:break_even_k} validates this approximation in wall-clock terms. We measure the end-to-end latency of \lightgluelate{$4$} (\lglate{4}) relative to \lglite{4}, a reduced-capacity baseline with the same online E-GF (defined in Sec.~\ref{sec:exp_setup}), on the Aachen dataset. At $K{=}1$, \lglate{4} is $1.59\times$ slower because its Context Block pass is amortized over a single database image. The ratio falls to $1.06\times$ by $K{=}10$ and converges toward parity thereafter, so at the $K \ge 50$ operating point of standard retrieval pipelines the one-time Context Block cost is effectively free.

Since standard image-matching baselines are evaluated only on isolated image pairs, E-GF provides a practical way to project their actual computational footprint in full visual localization pipelines. E-GF directly translates to end-to-end visual localization compute as demonstrated in Sec.~\ref{sec:results}.

\subsection{Entropy-guided Spatial Token Reduction and Expansion (\estoren)}\label{sec:estore}

\begin{wrapfigure}{r}{0.55\textwidth} 
    \vspace{-1em}
    \centering
    \includegraphics[width=\linewidth]{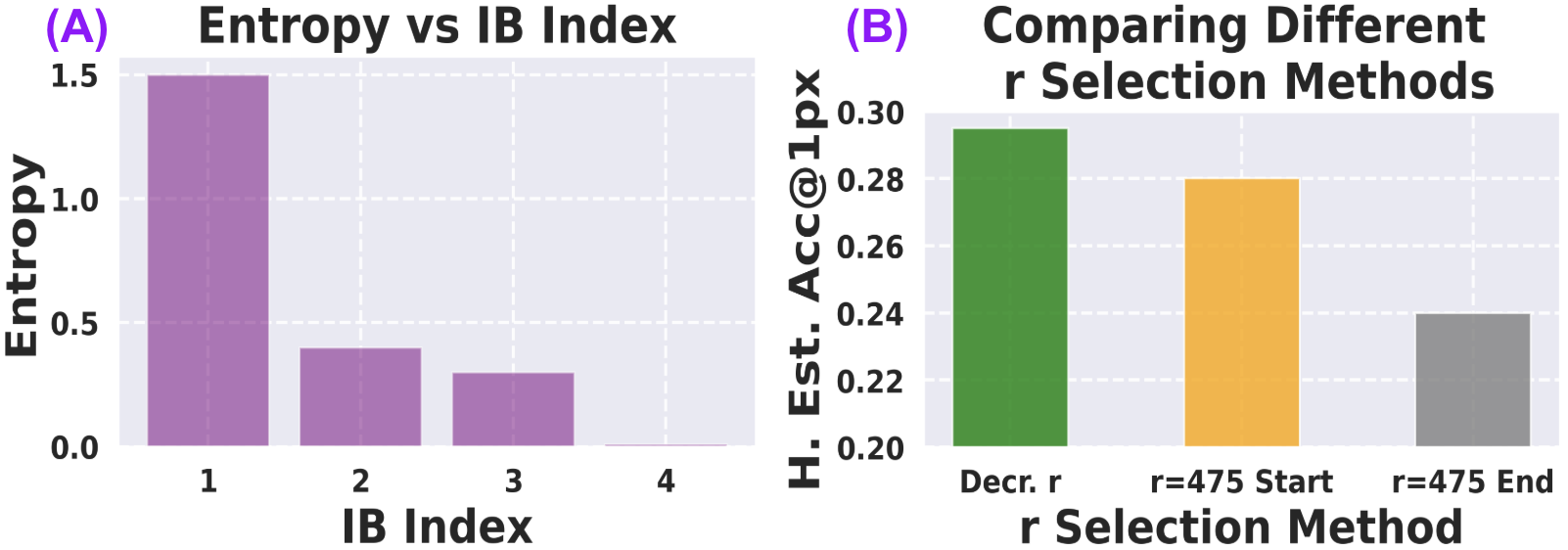} 
    \caption{
    \small
    \textbf{Cross-attention entropy naturally decays across Interaction Blocks, guiding the progressive token reduction in \estore.} \textbf{(A)} The cross-attention entropy across the four IBs of \lightgluelate[4]. Attention smoothly shifts from diffuse global consensus (high entropy) in early IBs to highly peaked, localized matching (low entropy) in later IBs. 
    \textbf{(B)} 
    Progressively decaying $r$ has significantly better performance (homography estimation accuracy) compared to other $r$ selection approaches.
    }
    \label{fig:entropy}
\end{wrapfigure}

While \coolname significantly reduces compute, we further optimize it by analyzing the interactions from an information-theoretic perspective. Let $\alpha_{i,j}$ be the cross-attention weight assigned by query token $i \in \{1, \dots, N_q\}$ to database token $j \in \{1, \dots, N_{db}\}$. The mean Shannon entropy $H$ of the entire cross-attention map is calculated by averaging the entropy across all $N_q$ query tokens:
{
\small
\begin{equation}
H = -\frac{1}{N_q} \sum_{i=1}^{N_q} \sum_{j=1}^{N_{db}} \alpha_{i,j} \log \alpha_{i,j}.
\end{equation}
}

Empirically, we observe that the entropy $H$ is high in the initial Interaction Blocks and monotonically decreases in successive layers. A high entropy indicates that early interactions are diffuse, attending broadly to many keypoints to establish global consensus. Conversely, low entropy in the final layers indicates highly peaked, localized attention necessary for precise geometric matching.
To exploit this behavior, we introduce Entropy-guided Spatial Token Reduction and Expansion (\estore). \estore merges the $r$ spatially closest token pairs before cross-attention, shortening the sequence by $r$ tokens. It then computes the interaction and unmerges the tokens for the subsequent layer. The reduction budget $r$ is initialized with a high value in the early IBs and progressively decreases in the later IBs. 
Unlike standard methods \citep{bolya2022tome} that merge tokens based on computationally heavy feature similarity, \estore relies on simple spatial proximity, computed from the keypoint positions $P$ already available from the feature extractor (Sec.~\ref{sec:one_to_k}), achieving comparable results at a fraction of the cost. Furthermore, while traditional methods permanently fuse tokens, destroying fine-grained spatial details, our novel \textit{merge $\rightarrow$ compute $\rightarrow$ unmerge} sequence preserves the exact keypoint geometry required for precise final matching.
In Sec.~\ref{sec:results} we show the advantage of using \estore and a comparison with other token merging strategies.  A detailed algorithm for \estore is provided in Appendix~\ref{app:estore}.

In \autoref{fig:entropy}\textbf{A}, we use our \lightgluelate[4] model (5 CBs, 4 IBs) to plot the cross-attention entropy of the four Interaction Blocks (from earliest to latest). We observe that the entropy is high in the first IB and successively decreases in later IBs, verifying our claim. 
We then apply \estore to \lightgluelate[4] using three different $r$ selection methods that yield identical compute reductions: applying a fixed $r=475$ only to the first IB, applying it to the last IB, and applying $r$ with a progressive decay across all four IBs ( $r = \{175, 150, 100, 50\}$). 
Applying $r=475$ solely to the last IB results in the worst performance, followed by applying reduction only to the first IB, while the progressive decay strategy performs the best (\autoref{fig:entropy}\textbf{B}). In all our experiments, we set $r$ to condense approximately 15\% of the keypoints in the first IB and reduce it to 10\% in the final IB.

\definecolor{sheetColor1}{HTML}{fff2cc}
\definecolor{sheetColor2}{HTML}{d9ead3}
\definecolor{sheetColor3}{HTML}{efefef}
\definecolor{sheetColor4}{HTML}{a4c5e0}


\begin{table*}[!htbp]
\centering
\begin{adjustbox}{max width=\textwidth}
\begin{NiceTabular}{c l l c l l c c l l c c l l c c l l}[colortbl-like]
\toprule
   & Method &  & \Block[c]{1-3}{\textbf{HPatches}} &  &  & \textbf{} & \Block[c]{1-3}{\textbf{ScanNet}} &  &  & \textbf{} & \Block[c]{1-3}{\textbf{MegaDepth}} &  &  & \textbf{} & \Block[c]{1-3}{\textbf{ETH-3D}} &  &  \\
\midrule
   &  &  & \Block[c]{1-3}{\makecell{Homography  \\ Est. Acc.}} &  &  &  & \Block[c]{1-3}{\makecell{Relative Pose  \\ Est. AUC}} &  &  &  & \Block[c]{1-3}{\makecell{Relative Pose  \\ Est. AUC}} &  &  &  & \Block[c]{1-3}{\makecell{Average Precision  \\ (AP)}} &  &  \\
   &  &  & 1px & 3px & E-GF &  & @$10^{\circ}$ & @$20^{\circ}$ & E-GF &  & @$10^{\circ}$ & @$20^{\circ}$ & E-GF &  & AP &  & E-GF \\
   &  &  &  &  &  &  &  &  &  &  &  &  &  &  &  &  &  \\
  \Block[fill=sheetColor1,c]{13-1}{\rotatebox{90}{LightGlue}} & LG-lite1 &  & 0.0691 & 0.2298 & 6.7 &  & 0.2112 & 0.3388 & 16.6 &  & 0.5797 & 0.7116 & 16.6 &  & 41.5 &  & 16.6 \\
   & \cellcolor{sheetColor2}LG-late1 (ours) & \cellcolor{sheetColor2} & \cellcolor{sheetColor2}\textbf{0.1334} & \cellcolor{sheetColor2}\textbf{0.3458} & \cellcolor{sheetColor2}6.7 & \cellcolor{sheetColor2}\textbf{} & \cellcolor{sheetColor2}\textbf{0.2557} & \cellcolor{sheetColor2}\textbf{0.3972} & \cellcolor{sheetColor2}16.6 & \cellcolor{sheetColor2}\textbf{} & \cellcolor{sheetColor2}\textbf{0.601} & \cellcolor{sheetColor2}\textbf{0.7374} & \cellcolor{sheetColor2}16.6 & \cellcolor{sheetColor2} & \cellcolor{sheetColor2}\textbf{56.24} & \cellcolor{sheetColor2} & \cellcolor{sheetColor2}16.6 \\
   &  &  &  &  &  &  &  &  &  &  &  &  &  &  &  &  &  \\
   & LG-lite2 &  & 0.1926 & 0.4653 & 13.4 &  & 0.2253 & 0.3633 & 33.3 &  & 0.6007 & 0.7352 & 33.3 &  & 49.16 &  & 33.3 \\
   & \cellcolor{sheetColor2}LG-late2 (ours) & \cellcolor{sheetColor2} & \cellcolor{sheetColor2}\textbf{0.2386} & \cellcolor{sheetColor2}\textbf{0.538} & \cellcolor{sheetColor2}13.4 & \cellcolor{sheetColor2}\textbf{} & \cellcolor{sheetColor2}\textbf{0.2784} & \cellcolor{sheetColor2}\textbf{0.4323} & \cellcolor{sheetColor2}33.3 & \cellcolor{sheetColor2}\textbf{} & \cellcolor{sheetColor2}\textbf{0.6354} & \cellcolor{sheetColor2}\textbf{0.7739} & \cellcolor{sheetColor2}33.3 & \cellcolor{sheetColor2} & \cellcolor{sheetColor2}\textbf{63.85} & \cellcolor{sheetColor2} & \cellcolor{sheetColor2}33.3 \\
   &  &  &  &  &  &  &  &  &  &  &  &  &  &  &  &  &  \\
   & LG-lite3 &  & 0.2417 & 0.5407 & 20.1 &  & 0.2573 & 0.3957 & 49.9 &  & 0.6096 & 0.7477 & 49.9 &  & 57.22 &  & 49.9 \\
   & \cellcolor{sheetColor2}LG-late3 (ours) & \cellcolor{sheetColor2} & \cellcolor{sheetColor2}\textbf{0.2859} & \cellcolor{sheetColor2}\textbf{0.5988} & \cellcolor{sheetColor2}20.1 & \cellcolor{sheetColor2}\textbf{} & \cellcolor{sheetColor2}\textbf{0.2898} & \cellcolor{sheetColor2}\textbf{0.4496} & \cellcolor{sheetColor2}49.9 & \cellcolor{sheetColor2}\textbf{} & \cellcolor{sheetColor2}\textbf{0.6272} & \cellcolor{sheetColor2}\textbf{0.7656} & \cellcolor{sheetColor2}49.9 & \cellcolor{sheetColor2} & \cellcolor{sheetColor2}\textbf{69.1} & \cellcolor{sheetColor2} & \cellcolor{sheetColor2}49.9 \\
   &  &  &  &  &  &  &  &  &  &  &  &  &  &  &  &  &  \\
   & LG-lite4 &  & 0.2681 & 0.5781 & 26.8 &  & 0.2638 & 0.416 & 66.6 &  & 0.6271 & 0.76 & 66.6 &  & 62.31 &  & 66.6 \\
   & \cellcolor{sheetColor2}LG-late4 (ours) & \cellcolor{sheetColor2} & \cellcolor{sheetColor2}\textbf{0.2957} & \cellcolor{sheetColor2}\textbf{0.6061} & \cellcolor{sheetColor2}26.8 & \cellcolor{sheetColor2}\textbf{} & \cellcolor{sheetColor2}\textbf{0.2988} & \cellcolor{sheetColor2}\textbf{0.4538} & \cellcolor{sheetColor2}66.6 & \cellcolor{sheetColor2}\textbf{} & \cellcolor{sheetColor2}\textbf{0.6392} & \cellcolor{sheetColor2}\textbf{0.7713} & \cellcolor{sheetColor2}66.6 & \cellcolor{sheetColor2} & \cellcolor{sheetColor2}\textbf{71.56} & \cellcolor{sheetColor2} & \cellcolor{sheetColor2}66.6 \\
   &  &  &  &  &  &  &  &  &  &  &  &  &  &  &  &  &  \\
   & \cellcolor{sheetColor3}\makecell{LG-default \\ (9 IBs)} & \cellcolor{sheetColor3} & \cellcolor{sheetColor3}0.3337 & \cellcolor{sheetColor3}0.6537 & \cellcolor{sheetColor3}60.4 & \cellcolor{sheetColor3} & \cellcolor{sheetColor3}0.3285 & \cellcolor{sheetColor3}0.4873 & \cellcolor{sheetColor3}149.9 & \cellcolor{sheetColor3} & \cellcolor{sheetColor3}0.6525 & \cellcolor{sheetColor3}0.7879 & \cellcolor{sheetColor3}149.9 & \cellcolor{sheetColor3} & \cellcolor{sheetColor3}75.14 & \cellcolor{sheetColor3} & \cellcolor{sheetColor3}149.9 \\
   &  &  &  &  &  &  &  &  &  &  &  &  &  &  &  &  &  \\
  \Block[fill=sheetColor4,c]{7-1}{\rotatebox{90}{MambaGlue}} & MG-lite3 &  & 0.2717 & 0.5712 & 33.2 &  & 0.2489 & 0.391 & 89.1 &  & 0.7175 & 0.8128 & 89.1 &  & 62.53 &  & 89.1 \\
   & \cellcolor{sheetColor2}MG-late3 (ours) & \cellcolor{sheetColor2} & \cellcolor{sheetColor2}\textbf{0.3101} & \cellcolor{sheetColor2}\textbf{0.618} & \cellcolor{sheetColor2}33.2 & \cellcolor{sheetColor2}\textbf{} & \cellcolor{sheetColor2}\textbf{0.2768} & \cellcolor{sheetColor2}\textbf{0.4308} & \cellcolor{sheetColor2}89.1 & \cellcolor{sheetColor2}\textbf{} & \cellcolor{sheetColor2}\textbf{0.7312} & \cellcolor{sheetColor2}\textbf{0.8322} & \cellcolor{sheetColor2}89.1 & \cellcolor{sheetColor2} & \cellcolor{sheetColor2}\textbf{69.39} & \cellcolor{sheetColor2} & \cellcolor{sheetColor2}89.1 \\
   &  &  &  &  &  &  &  &  &  &  &  &  &  &  &  &  &  \\
   & MG-lite4 &  & 0.3054 & 0.613 & 44.3 &  & 0.2726 & 0.4152 & 118.8 &  & 0.7266 & 0.8259 & 118.8 &  & 67.24 &  & 118.8 \\
   & \cellcolor{sheetColor2}MG-late4 (ours) & \cellcolor{sheetColor2} & \cellcolor{sheetColor2}\textbf{0.3334} & \cellcolor{sheetColor2}\textbf{0.6395} & \cellcolor{sheetColor2}44.3 & \cellcolor{sheetColor2}\textbf{} & \cellcolor{sheetColor2}\textbf{0.2909} & \cellcolor{sheetColor2}\textbf{0.4484} & \cellcolor{sheetColor2}118.8 & \cellcolor{sheetColor2}\textbf{} & \cellcolor{sheetColor2}\textbf{0.7309} & \cellcolor{sheetColor2}\textbf{0.8331} & \cellcolor{sheetColor2}118.8 & \cellcolor{sheetColor2} & \cellcolor{sheetColor2}\textbf{72.95} & \cellcolor{sheetColor2} & \cellcolor{sheetColor2}118.8 \\
   &  &  &  &  &  &  &  &  &  &  &  &  &  &  &  &  &  \\
   & \cellcolor{sheetColor3}\makecell{MG-default \\ (9 IBs)} & \cellcolor{sheetColor3} & \cellcolor{sheetColor3}0.3424 & \cellcolor{sheetColor3}0.6531 & \cellcolor{sheetColor3}99.7 & \cellcolor{sheetColor3} & \cellcolor{sheetColor3}0.3074 & \cellcolor{sheetColor3}0.465 & \cellcolor{sheetColor3}267.3 & \cellcolor{sheetColor3} & \cellcolor{sheetColor3}0.7379 & \cellcolor{sheetColor3}0.839 & \cellcolor{sheetColor3}267.3 & \cellcolor{sheetColor3} & \cellcolor{sheetColor3}74.75 & \cellcolor{sheetColor3} & \cellcolor{sheetColor3}267.3 \\
\bottomrule
\end{NiceTabular}
\end{adjustbox}
\caption{
    \small 
    \textbf{\coolname consistently and significantly outperforms computationally equivalent \emph{-lite} baselines.} We evaluate the \emph{-late} paradigm applied to \lightglue (top, yellow) and \mambaglue (bottom, blue) against equivalent reduced-capacity (\emph{-lite}) models. At identical online compute budgets (E-GFLOPs), structurally retaining deep intra-image context yields substantial improvements in matching accuracy over simply scaling down the network, with the most dramatic gains occurring under extreme computational constraints.
}
\label{tab:late}
\end{table*}

\vspace{-1em}
\section{Experiments}\label{sec:exp}
\subsection{Experimental Setup}\label{sec:exp_setup}

\textbf{Datasets.} We evaluate our method across five standard benchmarks that are widely used ~\citep{lightglue2023, superglue, ryoo2025mambaglue} to evaluate image-matching approaches. \emph{HPatches}~\citep{balntas2017hpatches} contains image pairs with known homography under varying illumination and viewpoint changes. We report homography estimation accuracy for different thresholds.
\emph{ScanNet}~\cite{dai2017scannet} contains images from indoor environments. Following standard practice~\citep{lightglue2023, ryoo2025mambaglue, loftr, gleize2023silk}, we report the area under the curve (AUC) of relative pose estimation accuracy at $10^{\circ}$ and $20^{\circ}$ angular thresholds. \emph{MegaDepth}~\citep{MegaDepthLi18} is an outdoor dataset where we report the relative pose estimation accuracy AUC at $10^{\circ}$ and $20^{\circ}$ thresholds. \emph{ETH-3D}~\cite{eth3d} benchmarks feature matching in both indoor and outdoor scans, where we report average precision (AP). \emph{Aachen Day-Night}~\citep{sattler2018aachen} is an end-to-end visual localization benchmark with severe day-to-night illumination changes, where we report pose recall at the $(0.5\,\text{m}, 5^\circ)$ threshold on the Day and Night splits.

\textbf{Baselines \& Notations.} 
In our \coolname paradigm, a model like \lightgluelate{$4$}, which utilizes 4 late Interaction Blocks and 5 Context Blocks, has the same effective compute (E-GFLOPs) as a standard \lightglue model with only 4 layers. Hence, we introduce a family of reduced-capacity ``\emph{-lite}'' models for both \lightglue and \mambaglue to serve as direct baselines for our ``\emph{-late}'' models. Similar to our ``\emph{-late}'' model notation (Sec.~\ref{sec:model_ins}), we use \lightgluelite{$x$} to represent a reduced-capacity \lightglue model with exactly $x$ layers, and \mambagluelite{$x$} to represent a reduced-capacity \mambaglue model with $x$ layers. Comparing a \emph{-late} model against its respective \emph{-lite} counterpart of equivalent effective cost directly demonstrates the advantage of late interaction. 
While \emph{-lite} models serve as our primary baseline, our paradigm is fully complementary to other efficiency strategies like token and model pruning, as shown in Appendix \ref{app:token_model_prune}. Additional implementation details are in Appendix \ref{app:imp_details}.

\begin{figure} 
    \centering
    \includegraphics[width=0.9\linewidth]{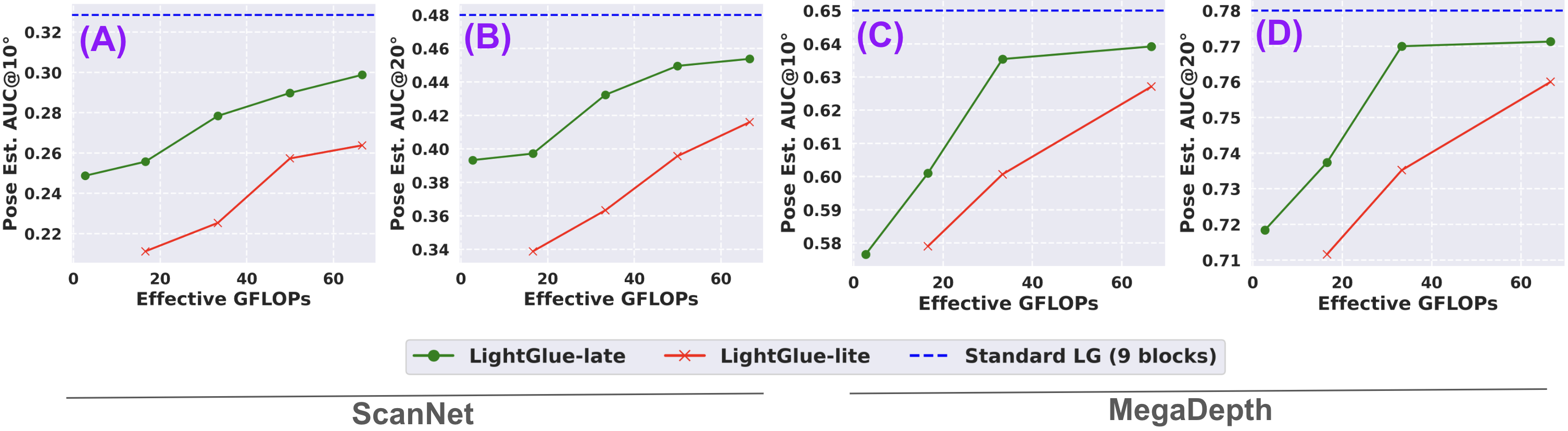} 
    \caption{
    \small
    \textbf{\coolname consistently dominates the compute-accuracy Pareto frontier.} Performance versus online computational cost (E-GFLOPs) evaluated on the ScanNet \textbf{(A, B)} and MegaDepth \textbf{(C, D)} datasets. Across all capacity levels, the \emph{-late} models (green) achieve strictly higher matching accuracy than the equivalent \emph{-lite} baselines (red) at the same computational budget.}
    \label{fig:qualitative}
    \vspace{-1em}
\end{figure}

\vspace{-1em}

\subsection{Results}
\label{sec:results}

\textbf{Primary Results:} \autoref{tab:late} presents our primary evaluation across the four image-matching datasets. We show the efficacy of the \coolname paradigm by applying it to both \lightglue (top, in yellow) and \mambaglue (bottom, in blue). For a rigorous comparison, we evaluate various capacities of our \emph{-late} models against their computationally equivalent \emph{-lite} counterparts, reporting multiple performance metrics and the effective online compute cost (E-GFLOPs) for each configuration.

Our results show that the \lightgluelate family consistently and significantly outperforms the \lightgluelite family across all model sizes and datasets. On average, the \coolname architecture yields a 14.94\% relative improvement in matching accuracy over the reduced-capacity \emph{-lite} baselines.  Crucially, the improvements are most pronounced under extreme computational constraints: \lightgluelate[1] achieves a substantial 28.84\%  average improvement over \lightgluelite[1]. As model capacity increases, \lightgluelate[4] maintains a solid 7.17\% improvement over \lightgluelite[4], achieving performance that closely rivals the full 9-layer model at over $2\times$ lower compute. We visualize this trend in \autoref{fig:qualitative} for MegaDepth and ScanNet.

We observe a similarly strong trend when applying \coolname to \mambaglue. While standard \mambaglue has a higher baseline performance than \lightglue, the \coolname restructuring still provides consistent, meaningful gains.
The \mambagluelate models consistently outperform their \mambagluelite equivalents, with \mambagluelate[3] and \mambagluelate[4] yielding average improvements of 7.24\% and 4.29\% over \mambagluelite[3] and \mambagluelite[4], respectively. 


\colorlet{darkgreen}{green!55!black}

\begin{table*}[!htbp]
\centering
\begin{adjustbox}{max width=\textwidth}
\begin{NiceTabular}{c c l | c l | c l}[colortbl-like]
\toprule
  \textbf{Dataset} & \Block[c]{1-2}{\textbf{LG-late4}} &  & \Block[c]{1-2}{\textbf{LG-late4  (ToMe + Unmerging)}} &  & \Block[c]{1-2}{\textbf{LG-late4 (ESToRE) - Ours}} &  \\
\midrule
  \textbf{} & \textbf{Perf.} & \textbf{E-GF} & \textbf{Perf.} & \textbf{E-GF} & \textbf{Perf.} & \textbf{E-GF} \\
   &  &  &  &  &  &  \\
  \textbf{\makecell{HPatches \\ (Acc@5px)}} & 0.6668 & 26.86 & 0.6567 (\textcolor{red}{$\downarrow 1.5\%$ }) & 25.15 (\textcolor{darkgreen}{$\downarrow 6.3\%$ }) & 0.6652 (\textcolor{red}{$\downarrow 0.23\%$ }) & 23.96 (\textcolor{darkgreen}{$\downarrow 10.79\%$ }) \\
  \textbf{\makecell{ScanNet \\ (AUC@10)}} & 0.2988 & 66.7 & 0.298 (\textcolor{red}{$\downarrow 0.26\%$ }) & 62.09 (\textcolor{darkgreen}{$\downarrow 6.9\%$ }) & 0.2969 (\textcolor{red}{$\downarrow 0.6\%$ }) & 57.86 (\textcolor{darkgreen}{$\downarrow 13.25\%$ }) \\
  \textbf{\makecell{MegaDepth \\ (AUC@20)}} & 0.7713 & 66.7 & 0.77 (\textcolor{red}{$\downarrow 0.17\%$ }) & 62.96 (\textcolor{darkgreen}{$\downarrow 5.60\%$ }) & 0.7701 (\textcolor{red}{$\downarrow 0.15\%$ }) & 58.7 (\textcolor{darkgreen}{$\downarrow 11.99\%$ }) \\
\bottomrule
\end{NiceTabular}
\end{adjustbox}
\caption{\textbf{\estore further reduces computational cost with negligible impact on performance.} \estore is applied to \lightgluelate[4], against ToMe \citep{bolya2022tome} augmented with our \textit{merge $\rightarrow$ compute $\rightarrow$ unmerge} mechanism for an apples-to-apples comparison. Both preserve accuracy, but \estore yields up to $2\times$ the reduction in Effective GFLOPs (E-GF). All percentages are relative to the LG-late4 column: \textcolor{red}{red} is accuracy lost, \textcolor{darkgreen}{green} is compute saved.}
\label{tab:estore_vs_tome}
\end{table*}
\textbf{\estoren Results: } \autoref{tab:estore_vs_tome} integrates \estore into \lightgluelate[4] with a decaying reduction budget $r$ (Sec.~\ref{sec:estore}). On HPatches, ScanNet, and MegaDepth, \estore reduces compute by 10.8\% to 13.3\% for a performance drop of only 0.2\% to 0.6\%.
We also compare against Token Merging (ToMe) \citep{bolya2022tome}. Because standard ToMe fuses tokens permanently, destroying the spatial geometry that precise matching requires, we first augment it with our \textit{merge $\rightarrow$ compute $\rightarrow$ unmerge} mechanism to obtain an apples-to-apples baseline. ToMe then matches \estore in accuracy, but \estore reduces E-GF substantially more, because it replaces ToMe's expensive cosine-similarity clustering with lightweight 2D spatial distance. Within image matching, spatial proximity is therefore a near-zero-cost proxy for semantic similarity during token merging.

\definecolor{sheetColor2}{HTML}{d9ead3}
\definecolor{sheetColor3}{HTML}{efefef}

\begin{wraptable}[11]{r}{0.44\textwidth}
\vspace{-2em}
\centering
\footnotesize
\begin{adjustbox}{max width=0.44\textwidth}
\begin{NiceTabular}{l c c c}[colortbl-like]
\toprule
  \textbf{Method} & \makecell{\textbf{Perf.} \\ (Day/Night)} & \textbf{GF} & \makecell{\textbf{Lat.} \\ (ms)} \\
\midrule
  LG-lite2 & 88.7/80.6 & 6.6k & 379 \\
  \cellcolor{sheetColor2}LG-late2 (ours) & \cellcolor{sheetColor2}\textbf{90.4/84.3} & \cellcolor{sheetColor2}6.7k & \cellcolor{sheetColor2}382 \\
\addlinespace
  LG-lite4 & 92.6/85.3 & 13.2k & 713 \\
  \cellcolor{sheetColor2}LG-late4 (ours) & \cellcolor{sheetColor2}\textbf{94.9/90.6} & \cellcolor{sheetColor2}13.3k & \cellcolor{sheetColor2}721 \\
\addlinespace
  \cellcolor{sheetColor3}LG-default & \cellcolor{sheetColor3}96.0/90.6 & \cellcolor{sheetColor3}28.3k & \cellcolor{sheetColor3}1619 \\
\bottomrule
\end{NiceTabular}
\end{adjustbox}
\caption{
    \footnotesize
    \textbf{\coolname preserves end-to-end accuracy at half the latency.} Aachen Day-Night recall at $(0.5\,\text{m}, 5^\circ)$, $K{=}50$, desktop GPU. GF and latency are for the full pipeline.
}
\label{tab:e2e_loc}
\end{wraptable}

\textbf{End-to-End Visual Localization Results:} 
\coolname reduces the cost of the matching stage, and Sec.~\ref{sec: e-gflops} showed that its one-time query overhead amortizes away. 
We now verify that these savings translate to a deployed system, evaluating the complete pipeline, comprising feature extraction, matching against $K{=}50$ retrieved candidates, and PnP with RANSAC, on the Aachen Day-Night dataset.
\autoref{tab:e2e_loc} reports end-to-end pose recall. \lglate{4} outperforms \lglite{4} by $2.3$ and $5.3$ points on the Day and Night splits respectively, matches the full 9-layer model on Night, and remains within $1.1$ points on Day while running $2.2\times$ faster. Peak memory is unchanged across all variants, because the cached Context Block outputs are the same size as the descriptors these pipelines already store.


\section{Conclusion, Limitations, and Future Work}\label{sec:limit}
We introduce \coolname, a caching paradigm for transformer-based (and variants) matchers that reduces effective inference compute by over 2$\times$, together with \estore, an entropy-guided token merging scheme that yields a further 10\% to 14\% savings with negligible accuracy loss. On a full localization pipeline these savings translate to a $2.2\times$ wall-clock speedup at pose accuracy comparable to the full-capacity matcher. Together, these methods mark a foundational shift toward efficient, real-time visual localization. The paradigm does rely on pre-computed caches, which restricts it to known, pre-mapped environments and rules out online odometry in unknown settings. It is also limited to sparse matchers; extending deferred interaction to monolithic, end-to-end dense architectures is a promising direction for future work.


\clearpage
\acknowledgments{
This work was supported in part by National Science Foundation grant No. 2148186, by the Office of Naval Research (ONR) under Grant No. N00014-22-1-2254, and by the Amazon AI PhD Fellowship. Any opinions, findings, conclusions, or recommendations expressed in this material are those of the authors and do not necessarily reflect the views of these agencies. We also thank Dr. Po-han Li for his valuable feedback and discussions on the paper.

}


\bibliography{bib/main, bib/swarm, bib/visual_loc}  

\newpage
\appendix

\section{Appendix}

\subsection{Implementation Details}\label{app:imp_details}
Following standard practice \citep{lightglue2023, ryoo2025mambaglue}, we first train all models on synthetic homographies of real images~\citep{revisitop} for 25 epochs followed by fine-tuning on MegaDepth~\citep{MegaDepthLi18} for 50 epochs. All experiments are conducted with a batch size of 128 and a learning rate of $1 \times 10^{-4}$. Training is distributed across 4 NVIDIA L40S GPUs. Pre-training on synthetic homographies converges in approximately 7 hours, while MegaDepth fine-tuning takes about 30 hours. Settings are kept strictly consistent across all models and baselines. Early exits are used to obtain various different capacities of \textit{lite} and \textit{late} models.

\subsection{The \estoren Algorithm} \label{app:estore}
Traditional token merging methods either use K-means~\citep{kmeans} or graph-cut~\citep{graphcut} to iteratively cluster and average tokens based on feature similarity. To reduce this clustering overhead, approaches like ToMe \citep{bolya2022tome} use bipartite matching to quickly fuse the most similar token pairs. However, applying these techniques directly to local feature matching presents two critical bottlenecks:
\begin{enumerate}[leftmargin=*]
    \item \textbf{Destruction of Geometry:} Existing methods fuse tokens permanently. Because tokens in image matching correspond to spatial keypoints, permanent fusion destroys their precise coordinates, making the output useless for geometric matching.
    \item \textbf{Compute Overhead:} Calculating feature similarity (e.g., cosine similarity) to determine which tokens to merge is inherently expensive and significantly diminishes the intended computational gains.
\end{enumerate}

To address these issues, we introduce \estore (Entropy-guided Spatial Token Reduction and Expansion). Instead of relying on expensive feature comparisons, \estore clusters tokens based purely on their 2D spatial distance in the image plane. We empirically demonstrate that this lightweight operation achieves performance on par with feature-similarity clustering but at a fraction of the cost. Furthermore, \estore avoids permanent fusion by utilizing a novel \textit{merge $\rightarrow$ compute $\rightarrow$ unmerge} paradigm as shown in Algorithm \ref{alg:estore}. This allows the network to exploit the efficiency of reduced token counts during heavy cross-attention layers while fully preserving the original keypoint geometry for the subsequent layers. To maintain conceptual clarity, Algorithm \ref{alg:estore} illustrates the high-level logic of \estore.

\textbf{Primitives.} Algorithm \ref{alg:estore} is written in terms of five elementary operations. $\mathrm{BipartitePartition}$ splits the tokens of one image into two disjoint sets, $A$ at even indices and $B$ at odd indices. $\mathrm{L2Distance}$ returns the $|A| \times |B|$ matrix of Euclidean distances between the keypoint positions of the two sets. $\mathrm{ArgMinK}$ returns the indices of the $r$ smallest entries of that matrix, that is, the $r$ spatially closest pairs across the two sets. $\mathrm{AverageFusedTokens}$ replaces each of those $r$ pairs by its mean and leaves every unselected token untouched, so the sequence shortens by exactly $r$. $\mathrm{ReconstructShape}$ inverts the pairing, writing each fused token back to the positions of both tokens it was formed from and thereby restoring the original sequence length. $\mathrm{SelfAttn}$, $\mathrm{CrossAttn}$ and $\mathrm{FFN}$ are the standard components of the underlying matcher and are unchanged by \estore.

\begin{algorithm}[H]
    \caption{Interaction Block Layer ($\text{Layer}^l_{IB}$) with \estore, shown for a single query-database pair (database index $k$ omitted).}
    \label{alg:estore}
    \SetAlgoLined
    \LinesNumbered
    \SetNlSty{}{}{\,\vrule width 0.4pt height 10pt depth 5pt}
    \DontPrintSemicolon
    
    \KwIn{Query features $F^{l-1}_q$, Database features $F^{l-1}_{db}$, Query keypoints $P_q$, Database keypoints $P_{db}$, Reduction budget $r$}
    \KwOut{Updated Query features $F^{l}_q$, Updated Database features $F^{l}_{db}$}
    \vspace{1mm}\hrule\vspace{1mm}
    
    \tcc{\hfill Intra-image Processing Phase \hfill\null}
    \tcp {Self-Attention (Or Variants)}
    $\hat{F}^l_q \leftarrow \text{SelfAttn}(F^{l-1}_q, P_q)$\label{ln:sa_q}\;
    $\hat{F}^l_{db} \leftarrow \text{SelfAttn}(F^{l-1}_{db}, P_{db})$\label{ln:sa_db}\;
    \BlankLine
    \BlankLine
    
    \tcc{\hfill Spatial Token Merging Phase \hfill\null}
    \tcp{Returns reduced features and an unmerge mapping $\mathcal{M}$}
    $\tilde{F}^l_q, \mathcal{M}_q \leftarrow \text{SpatialMerge}(\hat{F}^l_q, P_q, r)$\label{ln:merge_q}\;
    $\tilde{F}^l_{db}, \mathcal{M}_{db} \leftarrow \text{SpatialMerge}(\hat{F}^l_{db}, P_{db}, r)$\label{ln:merge_db}\;
    \BlankLine
    \BlankLine
    
    \tcc{\hfill Inter-image Interaction Phase \hfill\null} 
    \tcp{Cross-Attention on reduced tokens}
    $\text{msg}_q, \text{msg}_{db} \leftarrow \text{CrossAttn}(\tilde{F}^l_q, \tilde{F}^l_{db})$\label{ln:cross}\;
    \BlankLine
    \tcp{FFN applied to reduced tokens (concatenated with CrossAttn output)}
    $\tilde{F}'_q \leftarrow \text{FFN}([\tilde{F}^l_q \parallel \text{msg}_q])$\label{ln:ffn_q}\;
    $\tilde{F}'_{db} \leftarrow \text{FFN}([\tilde{F}^l_{db} \parallel \text{msg}_{db}])$\label{ln:ffn_db}\;
    \BlankLine
    \BlankLine
    
    \tcc{\hfill Token Unmerging \& Residual Phase \hfill\null}
    \tcp{Unmerge FFN outputs and add to full-length tokens}
    $F^{l}_q \leftarrow \hat{F}^l_q + \text{Unmerge}(\tilde{F}'_q, \mathcal{M}_q)$\label{ln:un_q}\;
    $F^{l}_{db} \leftarrow \hat{F}^l_{db} + \text{Unmerge}(\tilde{F}'_{db}, \mathcal{M}_{db})$\label{ln:un_db}\;
    \BlankLine
    \BlankLine
    
    \Return{$F^{l}_q, F^{l}_{db}$}\label{ln:ret}\;
    
    \vspace{0.3cm}
    
    \SetKwFunction{FTokenMerge}{SpatialMerge}
    \SetKwProg{Fn}{Function}{:}{}
    \Fn{\FTokenMerge{$F, P, r$}}{\label{ln:fn_merge}
        $A, B \leftarrow \text{BipartitePartition}(F, P)$\label{ln:partition} \tcp*{Split into Evens and Odds}
        $\Delta \leftarrow \text{L2Distance}(P_A, P_B)$\label{ln:dist} \tcp*{Compute spatial distance}
        $\mathcal{M} \leftarrow \text{ArgMinK}(\Delta, r)$\label{ln:argmin} \tcp*{The $r$ closest pairs}
        $\tilde{F} \leftarrow \text{AverageFusedTokens}(A, B, \mathcal{M})$\label{ln:avg}\;
        \KwRet{$\tilde{F}, \mathcal{M}$}\;
    }
    
    \vspace{0.3cm}
    
    \SetKwFunction{FTokenUnMerge}{Unmerge}
    \Fn{\FTokenUnMerge{$\tilde{F}, \mathcal{M}$}}{\label{ln:fn_unmerge}
        $F_{\text{restored}} \leftarrow \text{ReconstructShape}(\tilde{F}, \mathcal{M})$\label{ln:reconstruct} \tcp*{Duplicate fused tokens to original indices}
        \KwRet{$F_{\text{restored}}$}\;
    }
\end{algorithm}

\textbf{Walkthrough.} An Interaction Block equipped with \estore proceeds in four phases. Lines~\ref{ln:sa_q}--\ref{ln:sa_db} apply intra-image self-attention to the query and database tokens independently, exactly as in an unmodified Interaction Block, and operate on full-length sequences. Lines~\ref{ln:merge_q}--\ref{ln:merge_db} call $\mathrm{SpatialMerge}$ on each image, which returns a sequence shortened by $r$ tokens together with the mapping $\mathcal{M}$ required to invert the merge. The expensive inter-image computation then runs entirely on these shortened sequences: line~\ref{ln:cross} computes cross-attention between them, and lines~\ref{ln:ffn_q}--\ref{ln:ffn_db} apply the feed-forward network to each reduced sequence concatenated with its incoming messages. Lines~\ref{ln:un_q}--\ref{ln:un_db} undo the reduction, scattering the processed features back to full length and adding them to the pre-merge tokens $\hat{F}^l$ as a residual, so that layer $l+1$ receives one representation per original keypoint. Line~\ref{ln:ret} returns the updated representations.

The two helpers make the merge explicit and exactly invertible. Within $\mathrm{SpatialMerge}$ (line~\ref{ln:fn_merge}), line~\ref{ln:partition} splits the tokens into the two sets $A$ and $B$, line~\ref{ln:dist} computes their pairwise keypoint distances, line~\ref{ln:argmin} selects the $r$ closest pairs as the mapping $\mathcal{M}$, and line~\ref{ln:avg} averages each selected pair. $\mathrm{Unmerge}$ (line~\ref{ln:fn_unmerge}) is the exact inverse: line~\ref{ln:reconstruct} copies each fused token back to the two positions it came from. Because the reduction is confined to lines~\ref{ln:cross}--\ref{ln:ffn_db}, and both the self-attention of lines~\ref{ln:sa_q}--\ref{ln:sa_db} and the residual of lines~\ref{ln:un_q}--\ref{ln:un_db} act on full-length sequences, no keypoint is ever permanently fused. This is what preserves the exact geometry required for the final correspondence estimate.

\subsection{Additional Results with \mambaglue} \label{app:more_mamba}
\begin{figure}[h]
    \centering
    \includegraphics[width=\linewidth]{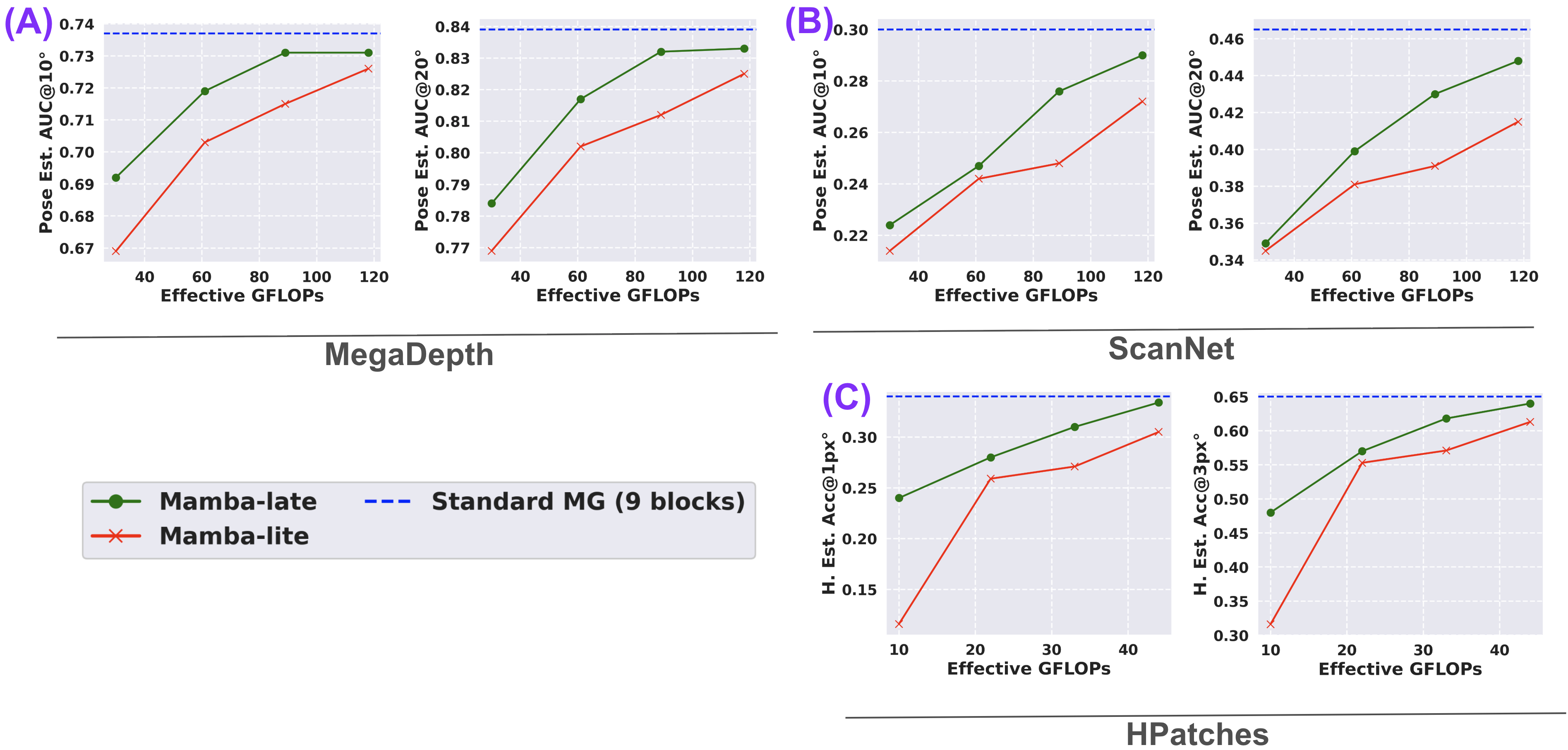}
    \caption{
    \small
    \textbf{\mambagluelate consistently and significantly outperforms computationally equivalent \mambagluelite baselines.} We show the trends on three datasets: \textbf{A)} MegaDepth, \textbf{B)} ScanNet, \textbf{C)} HPatches.}
    \label{fig:mamba_all_qual}
\end{figure}

In the main paper, we presented results for all different \lightgluelate models (\lightgluelate{1} to \lightgluelate{4}) but showed only \mambagluelate{3} and \mambagluelate{4} due to space constraints. In \autoref{fig:mamba_all_qual}, we show results with all four \mambagluelate models on MegaDepth, ScanNet, and HPatches. We see that the \emph{-late} variants consistently outperform the \emph{-lite} variants throughout the spectrum, and the results are more pronounced on the challenging MegaDepth and ScanNet datasets. 

\subsection{Token and Model Pruning} \label{app:token_model_prune}
\begin{figure}[h]
    \centering
    \includegraphics[width=0.7\linewidth]{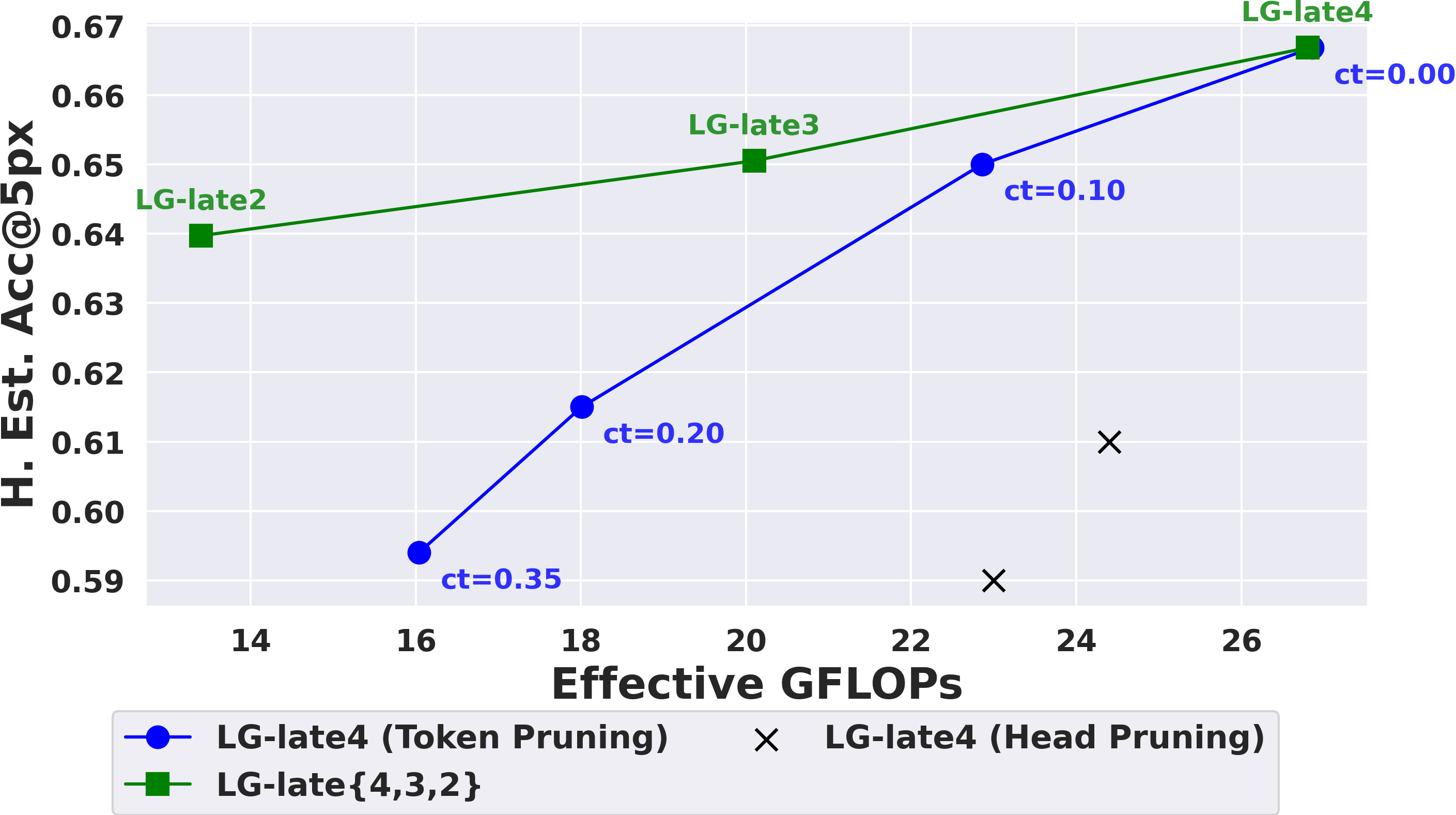}
    \caption{
    \small
    \textbf{\coolname outperforms token and model pruning.} Performance-compute trade-off on the HPatches dataset. We apply confidence-based token pruning, with different confidence thresholds ($ct$), to our \lightgluelate{4} model (blue) to progressively reduce its E-GFLOPs. We compare it against structurally smaller variants shown in green (\lightgluelate{2} and \lightgluelate{3}). We also compare against attention head pruning (black crosses). \coolname outperforms both these approaches by a significant margin. 
    } \label{fig:token_model_prune}
\end{figure}

In the main paper, we evaluated our approach against reduced-capacity \emph{-lite} models (Sec.~\ref{sec:exp_setup}) to establish a direct, apples-to-apples baseline. Here, we extend our evaluation by comparing the \coolname paradigm against other established inference speedup techniques, specifically token pruning and attention head pruning. For token pruning, we leverage the confidence scores predicted by \lightglue to retain only the most confident keypoints. Crucially, attempting to aggressively prune a standard \lightglue model to match the effective compute (E-GFLOPs) of \lightgluelate{4} required discarding an excessive number of tokens, which catastrophically degraded performance. Consequently, we apply token pruning directly on top of our \lightgluelate{4} model (using different confidence thresholds denoted by $ct$) and compare it against structurally smaller variants (\lightgluelate{2} and \lightgluelate{3}) on the HPatches dataset (Fig.~\ref{fig:token_model_prune}). The results demonstrate two key findings: first, \coolname yields a far superior performance-compute trade-off compared to aggressively dropping tokens; and second, our paradigm is fully complementary to existing pruning strategies, allowing them to be applied on top of it.
Furthermore, we explore sensitivity-based attention head pruning \citep{headprune} applied to \lightgluelate{4}. We find that pruning attention heads degrades performance significantly more than token pruning. This steep drop is likely due to the architectural constraints of \lightglue and \mambaglue, which utilize only $4$ attention heads, making the representations highly sensitive to the removal of any single head. 

Ultimately, these results prove that the \coolname paradigm offers a more robust foundational strategy for efficient image matching than standard pruning techniques, while simultaneously serving as an orthogonal framework that can seamlessly integrate with these traditional techniques.

\end{document}